\documentclass[11pt]{article}

\usepackage[preprint]{acl}
\usepackage{longtable}
\usepackage{times}
\usepackage{latexsym}
\usepackage{makecell} 
\usepackage{enumitem}
\usepackage{float}

\usepackage[T1]{fontenc}

\usepackage[utf8]{inputenc}

\usepackage{microtype}

\usepackage{inconsolata}

\usepackage{graphicx}

\title{SemiAdapt-Instruct: Extensible Instruction Tuning via Latent Domain-Specialised Adapters}

\author{Josh McGiff$^*$ \quad Salma Mekaoui \quad Robert Shanahan \quad Nikola S. Nikolov \\
  University of Limerick \\
  \texttt{josh.mcgiff@ul.ie} \\
  {\small $^*$Corresponding author}
}

\begin{document}
\maketitle
\begin{abstract}
Instruction-tuned LLMs are deployed into environments where domains evolve, yet extending a fine-tuned model's capabilities without full retraining remains an unsolved practical challenge. We present SemiAdapt-Instruct, a modular framework that discovers latent instruction domains, trains per-domain LoRA adapters in parallel, and performs parameter-free routing, incorporating new domains via single-adapter training without modifying existing components. SemiAdapt-Instruct outperforms full model fine-tuning across all configurations on both ROUGE-L and LLM-as-a-judge evaluation, while matching single LoRA fine-tuning and delivering extensibility that monolithic approaches cannot provide. We empirically demonstrate this extensibility by showing that updating a single adapter with new domain data outperforms all monolithic baselines. Our study 
also finds that independent discovery methods converge on the same specialisation-friendly domains. These findings demonstrate that decomposing 
heterogeneous instruction data into latent domains enables extensible NLP systems where evolving domains require only targeted single-adapter updates, eliminating the need for full model retraining.

\end{abstract}

\section{Introduction}

Supervised instruction fine-tuning has become the 
standard approach for adapting large language models 
to follow human instructions~\cite{wei2021finetuned, 
ouyang2022training}, yet the field has optimised for 
benchmark performance while leaving deployment 
maintainability largely 
unsolved~\cite{kiela-etal-2021-dynabench}. Models 
are evaluated on static benchmarks but deployed into 
environments where domains evolve and new 
capabilities are continuously required. However, full retraining on 
ever-growing combined datasets poses a structural barrier 
for practitioners without access to large-scale 
compute~\cite{mcgiff2025overcomingdatascarcitygenerative}.

Although parameter-efficient fine-tuning methods such as Low-Rank Adaptation (LoRA) reduce the computational cost of instruction tuning \cite{hu2021loralowrankadaptationlarge}, a single adapter trained on heterogeneous data inherits two problems. First, heterogeneous training data introduces 
conflicting gradient signals across 
domains~\cite{NEURIPS2020_3fe78a8a, ling2025domain}, 
which limits the degree to which a single model 
can specialise~\cite{NEURIPS2023_ec641387}. \begin{figure}[t]
\setlength{\tabcolsep}{6pt} 
  \includegraphics[width=\columnwidth]{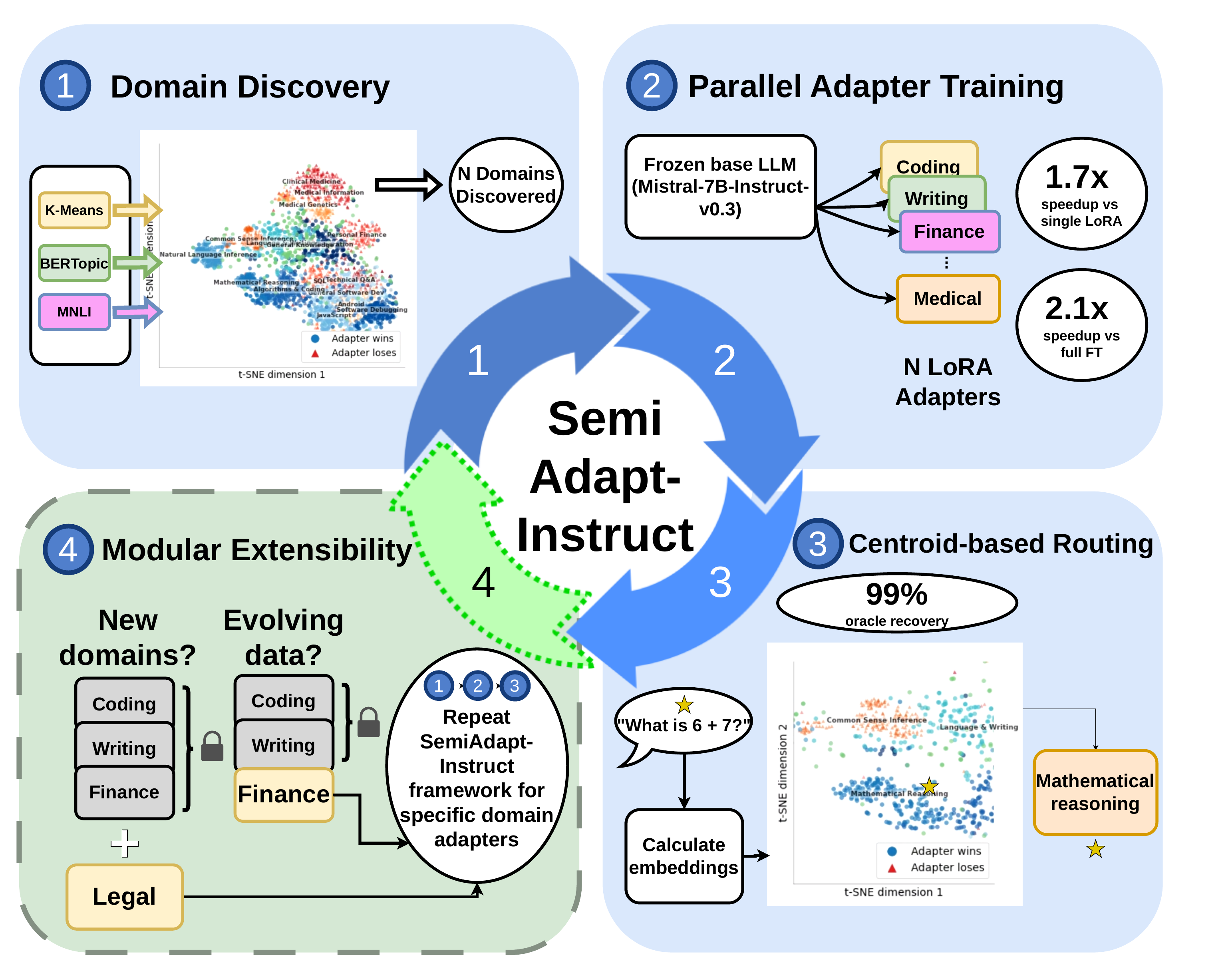}
  \caption{SemiAdapt-Instruct pipeline: (1) automatic 
domain discovery via unsupervised clustering or 
zero-shot classification, (2) independent parallel 
adapter training per domain, (3) parameter-free 
centroid routing at inference, and (4) modular 
extensibility via single-adapter updates for new 
domains or data.}
  \label{fig:kmeans-oracle}
\end{figure}Second, monolithic 
fine-tuning produces brittle, difficult-to-extend systems \cite{Huang_Yang_Wang_Qi_Yu_Fan_Wang_2026}. In such cases, incorporating new domains or updating existing knowledge
requires full retraining on the combined dataset, risking 
catastrophic forgetting of previously learned behaviours 
\cite{MCCLOSKEY1989109, li-etal-2024-revisiting, 11151751}. 

Standard Mixture-of-Experts (MoE) architectures address 
specialisation through jointly trained experts and 
routing components~\cite{shazeer2017outrageouslylargeneuralnetworks}, 
but do not resolve the extensibility problem. Joint 
training couples all experts together, meaning 
new domains cannot be added without retraining 
the full system. Wang et al.~\shortcite{NEURIPS2023_ec641387} 
confirm that instruction-tuned models on 
heterogeneous data fail to achieve consistent 
cross-domain performance, identifying modularity 
as a key unmet need. Parameter-efficient methods 
such as LoRA reduce the cost of retraining but 
do not eliminate it. Therefore, a single adapter trained 
on heterogeneous data must still be fully retrained 
to include new data.
\setlength{\parskip}{0pt}

Prior work~\cite{mcgiff2025semiadaptsemiloraefficientdomain} demonstrates that 
domain specialisation enables parameter-efficient 
methods to match or outperform full-model 
fine-tuning in a low-resource translation. However, that work focused on translation performance 
within a single fixed MNLI taxonomy and did not 
address extensibility, automatic domain discovery, 
or generalisation to heterogeneous instruction 
tuning.

\setlength{\parskip}{0pt}

In this work, we introduce SemiAdapt-Instruct, a modular framework that eliminates the retraining bottleneck for heterogeneous instruction tuning. Rather than assuming a fixed domain taxonomy, SemiAdapt-Instruct automatically discovers latent domains by comparing unsupervised clustering (K-means, BERTopic) and zero-shot classification (MNLI) across a dataset four times larger than prior work. Domain adapters are trained independently and in parallel with  routing occurring at inference via parameter-free cosine similarity to domain centroids. Unlike standard MoE or monolithic fine-tuning, new domains are incorporated by training a 
single additional adapter, while new data 
within an existing domain requires retraining 
only the relevant adapter, leaving all other 
components unchanged. Our contributions are as follows:

\begin{itemize}[noitemsep, topsep=-1pt, partopsep=0pt, itemsep=0pt]    
    
    \item \textbf{System.} SemiAdapt-Instruct, a modular 
instruction tuning framework that matches single LoRA 
fine-tuning and outperforms full model fine-tuning, 
while enabling domain updates that monolithic approaches 
cannot support.
    \item \textbf{Empirical Study.} A systematic 
    cross-method comparison of three automatic domain 
    discovery methods for instruction tuning.
    \item \textbf{Extensibility Demonstration.} 
A post-deployment domain update simulation 
showing that single-adapter retraining on new 
domain data outperforms monolithic 
baselines.
    \end{itemize}

\section{Related Work} 

\subsection{Instruction Tuning and Data Heterogeneity}

Instruction tuning has emerged as a fundamental post-training step for aligning large language models with human intent and expectations \cite{wei2021finetuned, ouyang2022training}. However, ever-expanding instruction datasets contribute to inherent heterogeneity. This presents a central challenge where training on dissimilar tasks introduces conflicting gradient signals and consequently degrades individual task performance \cite{NEURIPS2023_ec641387}.

Recent work addresses instruction data 
efficiency through filtering and quality 
valuation methods rather than structural 
partitioning. Centroid-based clustering with confidence-guided selection has been used to curate high-quality subsets for efficient training of instruction tuning models \cite{cai-etal-2025-low}. Similarly, ROSE~\cite{wu-etal-2025-rose} leverages 
pairwise preference loss as a reward signal 
to select high-utility samples for 
task-specific instruction tuning, while LimaCost~\cite{moon-etal-2025-limacost} 
estimates sample value by measuring 
gradient proximity to the LIMA dataset, 
selecting data that maximises alignment 
performance. These approaches 
improve data efficiency but discard low-scored 
samples, risking reduced generalisation to 
out-of-distribution examples. SemiAdapt-Instruct 
takes an alternative approach, partitioning the 
full instruction dataset into domain-specific 
subsets without discarding any examples, with 
each assigned to exactly one domain adapter.

   \subsection{Parameter-Efficient Fine-Tuning and Adapter Routing}
Parameter-efficient fine-tuning (PEFT) methods reduce the cost of adapting large pre-trained models by updating only a small fraction of parameters. One such method, Low-Rank Adaptation  \cite{hu2021loralowrankadaptationlarge}, achieves competitive performance with full fine-tuning at significantly reduced memory cost by expressing weight updates as low-rank matrix decompositions. However, catastrophic forgetting is a well-documented challenge in heterogeneous instruction tuning settings, whereby updating model parameters on a new task or domain degrades prior capabilities \cite{MCCLOSKEY1989109, li-etal-2024-revisiting, 11151751}. Recent work addresses this challenge through constrained LoRA updates. CLoRA~\cite{lu-etal-2025-controlled} introduces 
orthogonal subspace regularisation to preserve 
general capabilities during sequential 
fine-tuning of large language models. Alternatively, SLIM \cite{han-etal-2025-slim} proposes dynamic routing between LoRA adapters and identity layers to suppress forgetting. Pletenev~\emph{et al.}~\cite{pletenev-etal-2025-much} 
demonstrate that incorporating new knowledge 
via LoRA risks degrading previously learned 
capabilities, with bias amplified when training 
data is skewed towards specific entities.

The proliferation of domain-specific adapters has motivated work on multi-adapter routing. Tian ~\emph{et al.} \cite{tian-etal-2025-adapters} propose training a dedicated selector adapter to route between domain and task-specific LoRA modules at inference time. Similarly, GLIDER~\cite{li-etal-2025-glider} constructs 
LLM-generated global routing vectors combined 
with local task vectors to select from a pool 
of PEFT experts. Unlike these approaches, which require either a trained routing component or LLM-generated routing signals, our approach extends parameter-free routing to the instruction tuning setting.

    \vspace{-0.5em}

\subsection{Domain Discovery Methods}

Unsupervised discovery of latent structure in text 
has a long history in NLP. Latent Dirichlet 
Allocation~\cite{blei2003latent} introduced 
probabilistic topic modelling, representing documents 
as mixtures of topics and topics as distributions 
over words. Neural approaches such as 
BERTopic~\cite{grootendorst2022bertopic} and 
Top2Vec~\cite{angelov2020top2vec} extend this by 
clustering dense sentence embeddings, producing 
more semantically coherent topics. K-Means clustering has been used to identify representative subsets of full instruction datasets \cite{yu2024diversifyconquerdiversitycentricdata}. Zero-shot classification via natural language 
inference enables domain assignment using 
predefined label categories without task-specific 
training~\cite{yin-etal-2019-benchmarking}. SemiAdapt-Instruct 
conducts, to our knowledge, the first systematic comparison of clustering, topic modelling, 
and zero-shot classification, under identical 
controlled conditions for instruction tuning 
domain discovery.

\section{Methodology}
This section outlines our approach for discovering latent domains in raw instruction tuning data, training domain-specific adapters and evaluating our centroid-based routing method.

    \subsection{Latent Domain Discovery}
    This section describes three different approaches for discovering latent domains in a heterogeneous instruction dataset.

 \subsubsection{Unsupervised Clustering (K-Means)}

This experiment uses K-means clustering to partition 
training data into latent domains. Sentence embeddings 
are generated for each input using the all-MiniLM-L6-v2 
model prior to clustering \cite{NEURIPS2020_3f5ee243}.

\paragraph{Selecting K.} Identifying an appropriate 
cluster count proceeds in two stages. To avoid the 
computational cost of running full K-means across the 
entire dataset for every candidate value, Mini-Batch 
K-Means is employed during this exploratory phase, 
using batch sizes of 4,096 and three random 
initialisations. Inertia values are recorded across 
$K \in \{1, \ldots, 50\}$, with the KneeLocator 
algorithm \citep{Arvai_kneed_2026} applied to the resulting 
elbow curve to mathematically identify 
the inflection point at $K=17$.

To validate this, silhouette and Calinski-Harabasz 
scores are computed for $K \in \{14, 15, 16, 17, 18, 
19, 20\}$ on a 20,000-sample subset. Silhouette score peaks at $K=17$ (0.029), 
while Calinski-Harabasz peaks earlier at 
$K=14$ (212.4). We select $K=17$ as the point where the elbow criterion and silhouette score agree on an 
optimum. The low absolute silhouette value 
reflects the inherently overlapping surface patterns 
of instruction data rather than poor cluster quality,
a dissociation confirmed empirically by centroid 
routing accuracy of 95.0\% (Section~\ref{sec:routing}). Full K-means clustering is then run on the complete 
dataset using $K=17$.

        \subsubsection{Topic Modelling (BERTopic)}
\label{sec:bertopic}
Traditional topic modelling methods such as 
Latent Dirichlet Allocation are primarily 
designed for large text corpora and perform 
poorly on short texts like individual 
prompts~\cite{qiang2020short}. We therefore 
adopt BERTopic~\cite{grootendorst2022bertopic}, 
which generates contextual embeddings and 
applies context-aware clustering to identify 
coherent topics, automatically determining 
the number of domains~\cite{robledo2022topic} 
and yielding 107 topics. Given the 
identification of outlier examples under this 
approach, we additionally train adapters on 
super-groupings ($N=24$), where coarser topic 
clusters are derived via intertopic distance 
merging with outliers excluded, enabling a 
direct comparison of topic granularity within 
the same discovery framework.

        \subsubsection{Zero-Shot Classification (MNLI)}
        In this experiment, we apply zero-shot domain classification to the combined instruction dataset using \texttt{facebook/bart-large-mnli} \cite{lewis-etal-2020-bart}, a BART-based model fine-tuned on the Multi-Genre Natural Language Inference (MNLI) corpus \cite{williams-etal-2018-broad}. For each instruction, the model performs textual entailment between the instruction text and a set of hypothesis strings of the form \textit{"This text is about [label]"}, returning a probability distribution over the candidate labels. The label with the highest entailment probability is assigned as the domain, along with a confidence score. 
        
        The candidate label set is drawn directly from the 24 super topic groups produced by the BERTopic experiment described in Section~\ref{sec:bertopic}. This design choice serves two purposes. First, it enables a controlled comparison between a distributional clustering approach (BERTopic) and a classification-based approach (MNLI) over an identical label space, isolating the effect of the domain discovery mechanism rather than the label ontology. Second, it reflects a realistic deployment scenario in which an existing topic taxonomy is used to label new data without requiring model retraining.

\subsection{Domain-Specific Adapter Training}

Following domain discovery, we train a dedicated 
LoRA~\cite{hu2021loralowrankadaptationlarge} 
adapter for each identified domain across each domain discovery method. LoRA 
constrains weight updates to a low-rank subspace, 
preserving general model capabilities while 
enabling domain-specific specialisation, a 
property we validate empirically through 
comparison with full fine-tuning in 
Section~\ref{sec:results}.

Adapters are trained independently on each domain 
split, entirely decoupling domain discovery, 
adapter training, and inference routing. This 
modularity allows new domains to be incorporated 
by training a single additional adapter without 
retraining existing components, preserving 
existing adapter performance by construction 
since the base model remains frozen throughout.

\paragraph{Baselines.} We fine-tune two baselines 
on the full training set: a single LoRA
adapter providing a domain-agnostic parameter-efficient 
baseline, and a full fine-tune baseline 
updating all model parameters to assess whether 
parameter efficiency is a convenience or a necessity. 
Hyperparameters for all training runs are reported 
in Section~\ref{sec:setup}.

\subsection{Centroid-Based Inference Routing}

Domain specialisation requires identifying the 
correct domain at test time without ground-truth 
labels. We employ a parameter-free routing 
mechanism that requires no additional learned 
components. For each discovered domain $k$, a 
centroid $\mathbf{c}_k$ is pre-computed as the 
mean of all training instruction embeddings 
assigned to that cluster:

\begin{equation}
    \mathbf{c}_k = \frac{1}{|D_k|} \sum_{x_i \in D_k} 
    \phi(x_i)
\end{equation}

\noindent where $D_k$ is the set of training 
instructions assigned to domain $k$ and $\phi(x_i)$ 
denotes the sentence embedding of instruction $x_i$. 
At inference, a test instruction $x$ is routed to 
the domain $\hat{k}$ whose centroid has the highest 
cosine similarity:

\begin{equation}
    \hat{k} = \arg\max_{k} \frac{\phi(x) \cdot 
    \mathbf{c}_k}{\|\phi(x)\| \, \|\mathbf{c}_k\|}
\end{equation}

\noindent The adapter corresponding to domain 
$\hat{k}$ is then applied to the frozen base model 
for generation. This approach adds no trainable 
parameters and requires only a single embedding 
forward pass per inference call. We evaluate two routing configurations: an oracle setting, where each test instance is routed to its ground-truth domain adapter, and a centroid setting, where instances are routed to their predicted domain adapters.

\section{Experimental Setup}
\label{sec:setup}

We construct a heterogeneous instruction dataset 
by combining eight publicly available 
instruction-following corpora spanning diverse 
domains, yielding 509,174 total examples split 
into 458,255 training and 50,918 test examples 
via a 90/10 partition applied consistently 
across all discovery methods.

\begin{table}[htbp]
\centering
\tiny
\begin{tabular}{lccc}
\hline
\textbf{Dataset} & \textbf{Domain} & 
\textbf{\#Samples} & \textbf{Avg Tokens} \\
\hline
Alpaca-cleaned          & General      & 51,760  & $\sim$120 \\
CodeAlpaca-20K          & Code         & 20,022  & $\sim$40  \\
WebGPT                  & QA/Retrieval & 18,994  & $\sim$169 \\
Databricks Dolly 15K    & General      & 15,011  & $\sim$129 \\
Finance-Alpaca          & Finance      & 68,912  & $\sim$162 \\
AlpaCare-MedInstruct    & Medicine     & 52,002  & $\sim$184 \\
FLAN (CoT)              & Reasoning    & 74,771  & $\sim$70  \\
StackOverflow Dialogues & Programming  & 207,702 & $\sim$130 \\
\hline
\textbf{Total}          &              & \textbf{509,174} & \\
\hline
\end{tabular}
\caption{Composition and statistics of the instruction-tuning dataset. Additional dataset information is described in Appendix~\ref{sec:datadescription}}
\label{table:dataset}
\end{table}

All adapters are trained on Mistral-7B-Instruct-v0.3 
\cite{jiang2023mistral7b} with LoRA rank $r$=16, 
$\alpha$=32, dropout=0.05, targeting query, key, 
value, and output projections. Training uses batch 
size 16, learning rate 2e-4 with AdamW, for 1 epoch 
per domain on a single NVIDIA A100 80GB GPU.

We evaluate using ROUGE-L~\cite{lin-2004-rouge} computed via the 
\texttt{rouge-score} package\footnote{\url{https://github.com/google-research/rouge}} as a 
universal lexical metric enabling consistent comparison 
across all domains and discovery methods. To validate automatic metric findings, we 
additionally conduct blind pairwise preference evaluation 
using GPT-4o as an independent judge~\cite{NEURIPS2023_91f18a12} across all domain discovery methods, following 
established LLM-as-a-judge methodology.

\section{Results and Discussion}

\subsection{Overall Performance}
\label{sec:results}
Table~\ref{tab:main_results} presents ROUGE-L 
scores across all domain discovery methods on 
matched evaluation splits. Across all four 
configurations, oracle and centroid-routed 
adapter systems consistently outperform full 
model fine-tuning ($\Delta$ = +0.013--+0.026 
ROUGE-L), demonstrating that modular 
parameter-efficient adaptation is preferable 
to monolithic full-model retraining on 
heterogeneous instruction data regardless of 
discovery method choice.

Relative to the more competitive single LoRA 
baseline, K-means (K=17) is the only method 
where both oracle and centroid routing exceed 
the matched single LoRA score, achieving 
ROUGE-L of 0.4341 and 0.4314 respectively 
against 0.4282 for single LoRA. BERTopic 
super-24 oracle also outperforms its matched 
single LoRA baseline (+0.003). Across all 
methods, oracle ROUGE-L differences relative 
to single LoRA remain within 0.01, confirming 
broad performance parity. GPT-4o pairwise 
evaluation further confirms this parity as 
K-means and BERTopic-107 achieve near-identical 
GPT-4o preference rates (48.8\% and 48.5\%), 
while MNLI and BERTopic super-24 show modest 
judge preference for single LoRA (41.0\% and 
43.0\%), consistent with their negative ROUGE-L 
deltas. Per-domain analysis revealing where 
specialisation gains concentrate is presented 
in Section~\ref{sec:perdomain}.

Notably, ROUGE-L and GPT-4o evaluation diverge 
for BERTopic-107. Despite a negative ROUGE-L 
delta ($-$0.006), GPT-4o judges its responses 
as comparable to single LoRA (48.5\%), 
suggesting that lexical overlap is an 
imperfect proxy for response quality in 
latent domain instruction evaluation.

Across all configurations, modular domain 
specialisation achieves performance parity 
with single LoRA fine-tuning while outperforming 
full model fine-tuning, with the added advantage 
that new domains and data require only 
single-adapter updates.

\begin{table*}
  \centering
  \scriptsize
  \begin{tabular}{lcccccccc}
    \hline
    \textbf{Method} & \textbf{Oracle} & \textbf{Centroid} & 
    \textbf{LoRA} & \textbf{Full-FT} & 
    \textbf{$\Delta$ (Ora$-$LoRA)} & 
    \textbf{$\Delta$ (Ora$-$FT)} &
    \textbf{GPT-4o vs LoRA} & 
    \textbf{GPT-4o vs FT} \\
    \hline
    K-means (K=17)   & \textbf{0.4341} & \textbf{0.4314} & 0.4282 & 0.4082 & $\mathbf{+0.0059}$ & $\mathbf{+0.0259}$ & \textbf{48.8\%} & \textbf{63.2\%} \\
    BERTopic (N=107) & 0.4371 & 0.4324 & 0.4435 & 0.4240 & $-0.0064$ & $+0.0131$ & 48.5\% & 58.0\% \\
    MNLI (N=24)      & 0.4166 & 0.4083 & 0.4169 & 0.3995 & $-0.0003$ & $+0.0171$ & 41.0\% & 60.5\% \\
    BERTopic (N=24)  & 0.4446 & 0.4369 & 0.4412 & 0.4220 & $+0.0030$ & $+0.0226$ & 43.0\% & 62.0\% \\
    \hline
  \end{tabular}
  \caption{ROUGE-L on matched evaluation splits 
  (500 examples per domain). Single LoRA and Full 
  FT scores computed on same matched subsets. 
  $\Delta$ (Ora$-$LoRA) = oracle minus single LoRA; 
  $\Delta$ (Ora$-$FT) = oracle minus full fine-tuning. 
  GPT-4o Win\% 
  computed on 200 pairs per method (1,700 for 
  K-means, see Section~\ref{sec:perdomain}).}
  \label{tab:main_results}
\end{table*}

\subsection{Per-Domain Analysis}
\label{sec:perdomain}

    Table~\ref{tab:kmeans_perdomain} and 
    Figure~\ref{fig:kmeans-overlap} in 
    Appendix~\ref{sec:extendanalysis} show that 
    9 of 17 K-means domains achieve positive 
    ROUGE-L deltas relative to single LoRA, with 
    winning domains averaging $+0.007$ and losing 
    domains averaging $-0.008$. GPT-4o pairwise evaluation 
    confirms parity with single LoRA (48.8\% 
    average win rate) while showing consistent 
    preference for domain adapters over full 
    fine-tuning across all 17 domains (51.0--74.0\%, 
    mean 63.2\%). Together, these results 
    demonstrate that domain specialisation matches 
    single LoRA quality while outperforming full 
    model fine-tuning, delivering an extensible 
    system without systematic quality degradation.

\begin{table*}
  \centering
  \scriptsize
\begin{tabular}{clccccccccc}
    \hline
    \textbf{Cl.} & \textbf{Domain} & \textbf{Train N} & 
    \textbf{Oracle} & \textbf{LoRA} & \textbf{FT} &
    \textbf{$\Delta$ (O$-$L)} &
    \textbf{$\Delta$ (O$-$FT)} &
    \textbf{vs LoRA} &
    \textbf{vs FT} \\
    \hline
    \hline
    c10 & Mathematical Reasoning  & 14,744 & 0.5249 & 0.5042 & 0.5019 & $+0.021$ & $+0.023$ & 44\% & 66\% \\
    c2  & Technical Q\&A          & 34,393 & 0.4063 & 0.3939 & 0.3608 & $+0.012$ & $+0.045$ & 43\% & 63\% \\
    c14 & General SW Dev          & 48,069 & 0.4428 & 0.4316 & 0.4141 & $+0.011$ & $+0.029$ & 48\% & 74\% \\
    c5  & Language \& Writing     & 33,865 & 0.4672 & 0.4594 & 0.4275 & $+0.008$ & $+0.040$ & 50\% & 74\% \\
    c11 & NL Inference            & 37,241 & 0.5909 & 0.5833 & 0.5789 & $+0.008$ & $+0.012$ & 51\% & 59\% \\
    c4  & Content Gen             & 32,297 & 0.3173 & 0.3144 & 0.2952 & $+0.003$ & $+0.022$ & 45\% & 62\% \\
    c15 & Algorithms \& Coding    & 24,790 & 0.4765 & 0.4738 & 0.4700 & $+0.003$ & $+0.007$ & 44\% & 64\% \\
    c13 & JavaScript/jQuery       & 28,916 & 0.4264 & 0.4241 & 0.4034 & $+0.002$ & $+0.023$ & 61\% & 62\% \\
    c8  & General Knowledge       & 48,487 & 0.4684 & 0.4667 & 0.4308 & $+0.002$ & $+0.038$ & 54\% & 62\% \\
    c12 & Common Sense Inf.       & 6,956  & 0.5591 & 0.5595 & 0.5524 & $-0.000$ & $+0.007$ & 47\% & 51\% \\
    c7  & Personal Finance        & 20,204 & 0.1956 & 0.1978 & 0.1928 & $-0.002$ & $+0.003$ & 45\% & 59\% \\
    c6  & Clinical Medicine       & 14,515 & 0.4337 & 0.4361 & 0.4182 & $-0.002$ & $+0.016$ & 58\% & 61\% \\
    c0  & Medical Info            & 20,929 & 0.4002 & 0.4039 & 0.3651 & $-0.004$ & $+0.035$ & 48\% & 57\% \\
    c16 & Medical Genetics        & 13,357 & 0.4058 & 0.4152 & 0.3739 & $-0.010$ & $+0.032$ & 53\% & 64\% \\
    c3  & SW Debugging            & 37,935 & 0.3823 & 0.3944 & 0.3719 & $-0.012$ & $+0.010$ & 43\% & 65\% \\
    c1  & Android/Mobile Dev      & 21,225 & 0.3972 & 0.4105 & 0.3868 & $-0.013$ & $+0.010$ & 46\% & 64\% \\
    c9  & SQL/Databases           & 20,332 & 0.4566 & 0.4702 & 0.4484 & $-0.014$ & $+0.008$ & 49\% & 67\% \\
    \hline
    & \textbf{Macro Avg} & & \textbf{0.4324} & 0.4282 & 0.4160 &
    $\mathbf{+0.004}$ & $\mathbf{+0.021}$ & \textbf{48.8\%} & \textbf{63.2\%} \\
    \hline
  \end{tabular}
  \caption{Per-domain oracle evaluation for K-means ($K$=17) 
  on matched 500-example evaluation subsets. 
  $\Delta$ (O$-$L) and $\Delta$ (O$-$FT) report oracle minus 
  single LoRA and oracle minus full fine-tuning ROUGE-L 
  respectively. The \textit{vs LoRA} and \textit{vs FT} columns report GPT-4o pairwise 
  preference proportions from 100 comparisons per domain. }
  \label{tab:kmeans_perdomain}
\end{table*}

From a ROUGE-L perspective, 
the largest gains concentrate in Mathematical 
Reasoning (c10, $\Delta=+0.021$), Technical 
Q\&A (c2, $\Delta=+0.012$), and General 
Software Development (c14, $\Delta=+0.011$). 
GPT-4o win rates versus single LoRA reveal 
that ROUGE-L gains do not uniformly reflect 
preference improvements: Mathematical 
Reasoning achieves only 44\% preference over 
single LoRA despite a positive ROUGE-L delta 
($+0.021$), suggesting adapters learn response 
structure without improving semantic 
correctness.

Several domains show agreement between 
ROUGE-L and GPT-4o evaluation. JavaScript/jQuery, General Knowledge and 
Natural Language Inference all show consistent positive 
signals across both metrics, suggesting domain 
specialisation provides reliable quality 
improvement in these domains. GPT-4o preference for domain adapters over 
full fine-tuning is consistent across the 
top-performing domains, with General Software 
Dev (74\%), Mathematical Reasoning (66\%), 
and Technical Q\&A (63\%) all showing 
substantial adapter preference.

The medical domains present an interesting 
metric divergence. Medical Genetics shows a 
meaningful ROUGE-L loss ($\Delta=-0.010$) yet 
GPT-4o preferences tell the opposite story, 
with Medical Genetics achieving 53\% and 
Clinical Medicine 58\% win rates despite 
neutral or negative lexical overlap 
($-0.010$ and $\Delta=-0.002$ respectively). 
This suggests domain-specialised adapters 
produce higher-quality medical responses that 
diverge lexically from reference answers, 
consistent with known limitations of lexical 
overlap metrics for paraphrase-rich 
text~\cite{zhang2020bertscoreevaluatingtextgeneration}. 
Domain specialisation in medical contexts may 
therefore be underestimated by ROUGE-L alone. 
GPT-4o preference over full fine-tuning 
remains positive for both domains, with 
Clinical Medicine at 61\% and Medical Genetics 
at 64\%.

Discovering balanced latent domain groupings 
appears to be a key factor in adapter 
performance. The three medical 
domains (Medical Genetics, Clinical Medicine, 
and Medical Information) occupy overlapping 
regions of the embedding space 
(Figure~\ref{fig:tsne} in 
Appendix~\ref{sec:appendix}), proximate to 
Language \& Writing, Content Generation, and 
General Knowledge clusters. This proximity 
suggests medical instruction following 
requires general language production skills 
distributed across multiple topical clusters 
rather than concentrated in domain-specific 
data. Representation sharing across these 
related clusters could benefit medical adapter 
performance, and motivates exploring 
functional domain discovery as a complement 
to topical clustering. That is, domain 
discovery may benefit from dimensions relating 
to the kind of task an instruction requires, 
not only its topic. Future work should compare 
latent topical and functional domain discovery 
for heterogeneous instruction tuning.

These results collectively demonstrate that 
evaluation of domain-specialised instruction 
tuning requires metrics sensitive to semantic 
equivalence, particularly in open-ended 
medical and general knowledge domains where 
lexical variation between valid responses is 
high. Notably, GPT-4o preference over full 
fine-tuning remains consistently positive 
across medical domains despite ROUGE-L losses, 
confirming that adapter quality in these 
domains is underestimated by lexical metrics 
alone.

\subsection{Cross-Method Domain Convergence}

\begin{figure}[t]
  \includegraphics[width=\columnwidth]{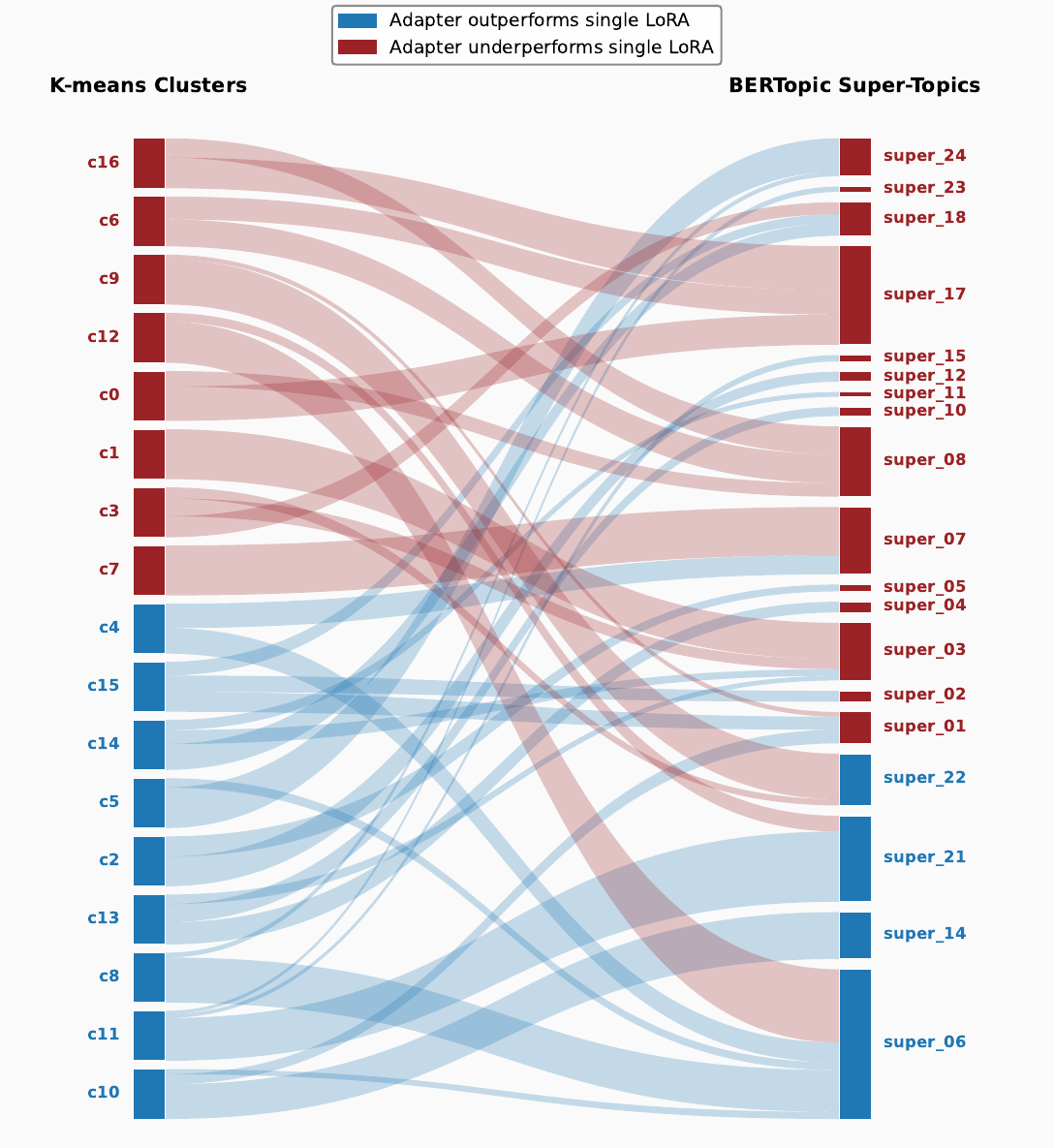}
  \caption{Cross-method domain alignment between 
  K-means clusters (left) and BERTopic super-24 
  topics (right). Flow thickness indicates the 
  proportion of examples shared between domains. 
  In terms of ROUGE-L, adapter-winning domains are shown in blue and adapter-losing domains in red.}

  \label{fig:shareddata}
\end{figure}

Figure~\ref{fig:shareddata} explores whether 
domain specialisation benefit is an intrinsic 
property of instruction content or an artefact 
of the discovery mechanism. We analyse the 
overlap between K-means winning domains and 
their corresponding BERTopic super-24 
groupings, with several consistent patterns emerging. 
Adapter-winning K-means clusters predominantly 
map to adapter-winning BERTopic super-groups: 
c10 maps 50.7\% of examples to super\_14 
(BERTopic WIN) and c11 maps 76.0\% to 
super\_21 (BERTopic WIN). Conversely, 
adapter-losing clusters predominantly map to 
adapter-losing super-groups, with c7 mapping 
52.2\% to super\_07 and c16 and c6 mapping 
predominantly to super\_17 and super\_08. 
This convergence across two structurally 
independent discovery methods suggests that 
specialisation benefit is largely determined 
by instruction content rather than the 
discovery mechanism itself.

To further validate this finding at a finer 
granularity, we map K-means cluster assignments 
to BERTopic-107 topic assignments for the same 
instructions. The K-means mathematical reasoning cluster maps 
predominantly to a winning BERTopic-107 topic 
(27.0\% of examples), while natural language 
inference and general knowledge clusters map to 
winning BERTopic-107 topics in their top 
mappings. Conversely, K-means losing domains 
such as personal finance map consistently to 
losing BERTopic-107 topics. Alignment is weaker 
for domains such as 
JavaScript/jQuery, where content is distributed 
across multiple fine-grained topics with mixed 
outcomes. Together, comparing K-means with both BERTopic methods futher indicates that semantic coherence 
of domain content determines specialisation 
benefit rather than the discovery mechanism 
itself.

However, two notable exceptions exist: c9 and 
c12 are adapter-losing K-means clusters whose 
examples map predominantly to BERTopic 
adapter-winning super-groups (super\_22 and 
super\_06 respectively, with 49.2\% and 79.2\% 
overlap). We investigate whether running 
BERTopic adapters on these K-means test 
examples recovers performance. 
Table~\ref{tab:hybrid_routing} shows that 
neither BERTopic adapter matches single LoRA 
on these examples. The c9 adapter improves 
marginally over the K-means adapter ($+0.004$) 
but remains below single LoRA, while c12 
performs worse than both. GPT-4o evaluation 
provides further context: c9 and c12 achieve 
near-parity with single LoRA under 
preference-based evaluation (49\% and 47\% 
respectively), suggesting that ROUGE-L 
underestimates adapter quality for these 
domains. Notably, c9 achieves 67\% GPT-4o 
preference over full fine-tuning, consistent 
with the broader pattern of adapter superiority 
over monolithic retraining. Common Sense 
Inference (c12) shows the weakest full 
fine-tuning preference at 51\%, which may 
reflect its dependence on broad distributional 
knowledge that monolithic training captures 
more effectively than domain-specific 
adaptation. Whether this reflects an intrinsic 
property of common sense reasoning or a 
limitation of the current dataset composition 
remains an open question.

\begin{table}
  \centering
  \tiny
  \begin{tabular}{clccc}
    \hline
    \textbf{Cluster} &  \textbf{Single} & 
    \textbf{K-means} & \textbf{BERTopic} \\
     & \textbf{LoRA} & & & \\
    \hline
    c9  & 0.4702 & 0.4615 & 0.4603 \\
    c12 & 0.5595 & 0.5517 & 0.5541 \\
    \hline
  \end{tabular}
  \caption{Cross-method adapter evaluation: BERTopic super-24 adapters 
  (super\_22 and super\_06 respectively) evaluated on K-means c9 and c12 
  test examples.}   
  \label{tab:hybrid_routing}
\end{table}

\subsection{Centroid Routing Retention}
\label{sec:routing}
Table~\ref{tab:routing_accuracy} reports centroid 
routing accuracy across all domain discovery methods. 
K-means achieves the highest accuracy at 95.0\%, 
with BERTopic-107 at 85.9\% and BERTopic super-24 
at 78.6\%. MNLI achieves the lowest accuracy at 
64.4\%, reflecting a fundamental limitation of 
zero-shot classification as a discovery mechanism. Imposing predefined categories rather than 
discovering latent distributional structure produces 
less discriminative centroid representations and 
correspondingly reduced GPT-4o preference (41.0\%). 

Despite lower routing accuracy than K-means, 
BERTopic-107 achieves comparable GPT-4o preference 
(48.5\% vs 48.8\%), suggesting routing errors do 
not proportionally degrade response quality. This is explained by the semantic structure 
of BERTopic-107, where intertopic distance 
analysis reveals substantial overlap between 
fine-grained topics, motivating the super-24 
groupings described in 
Section~\ref{sec:bertopic}. Routing errors 
predominantly represent assignments to semantically 
adjacent topics whose adapters are trained on 
overlapping content, such as misrouting between 
\textit{50\_jquery\_element\_div} and 
\textit{30\_div\_css\_height\_width}, rather than 
cross-domain failures. Quality is preserved because 
the assigned adapter remains domain-appropriate at 
a functional level. Parameter-free centroid routing therefore removes the need for a trained routing 
classifier, requiring only recomputation of 
the mean embedding of a new domain's training 
instructions and making domain extension 
lightweight by design.

\begin{table}
\tiny
  \centering
  \begin{tabular}{lcc}
    \hline
    \textbf{Method} & \textbf{Correct} & \textbf{Accuracy} \\
    \hline
    \textbf{K-means (K=17) }         & \textbf{8073/8500}   & $\mathbf{95.0\%}$ \\
    BERTopic (107 topics)   & 19962/23251 & 85.9\% \\
    BERTopic (super-24)     & 22771/28985 & 78.6\% \\
    MNLI (24 labels)        & 4929/7650   & 64.4\% \\
    \hline
  \end{tabular}
  \caption{Centroid routing accuracy across domain discovery methods.
  Accuracy measures the proportion of test examples assigned to the
  correct domain by cosine similarity to domain centroids, relative
  to oracle domain labels.}
  \label{tab:routing_accuracy}
\end{table}

\subsection{Modular Extensibility and Efficiency}

Domain extensibility is a structural property 
of the SemiAdapt-Instruct architecture. 
Existing domains are updated by retraining 
only the relevant adapter on new data, while 
new domains are incorporated by training a 
single additional adapter, in both cases 
leaving all other components unchanged. To empirically validate 
this claim, we simulate a post-deployment 
domain update. That is, a 10,000-example subset of 
Magicoder~\cite{wei2024magicoderempoweringcodegeneration} is assigned 
to K-means domains via existing centroid 
routing, with 4,199 examples mapped to the 
Algorithms \& Coding cluster (c15). Only this 
adapter is retrained on the combined original 
and new data, while all 16 remaining adapters are 
unchanged. The updated adapter achieves 
ROUGE-L of 0.553 and BERTScore F1 of 0.934 
on 200 held-out Magicoder examples, 
substantially outperforming all frozen 
baselines (Table~\ref{tab:extensibility}). 
GPT-4o pairwise evaluation confirms strong 
preference for the updated adapter over the 
original frozen adapter (65.0\%), single LoRA 
(66.5\%), and full fine-tuning (81.5\%). This demonstrates 
that targeted single-adapter updates 
incorporating new domain data outperform 
frozen monolithic baselines that cannot adapt 
without full retraining.

\begin{table}[h]
  \centering
  \tiny
  \begin{tabular}{lccc}
    \hline
    \textbf{Method} & \textbf{ROUGE-L} & 
    \textbf{BERTScore F1} & 
    \textbf{Updated Preferred} \\
    \hline
    \textbf{Updated adapter (c15)} & \textbf{0.553} & 
    \textbf{0.934} & --- \\
    Original adapter (c15) & 0.454 & 0.912 & 
    65.0\% \\
    Single LoRA             & 0.442 & 0.909 & 
    66.5\% \\
    Full Fine-Tuning        & 0.392 & 0.903 & 
    \textbf{81.5\%} \\
    \hline
  \end{tabular}
  \caption{Extensibility experiment: updating 
  only the Algorithms \& Coding adapter (c15) 
  with 4,199 new Magicoder examples while 
  leaving all other adapters unchanged. 
  \textit{Updated Preferred} reports the 
  proportion of 200 GPT-4o pairwise comparisons 
  where the updated adapter is preferred over 
  each frozen baseline.}
  \label{tab:extensibility}
\end{table}

Decomposing heterogeneous data into independent 
domains enables parallel adapter training, with each adapter training on its own data partition with no inter-adapter dependencies. In our experiments, 
three adapters were trained simultaneously on a 
single A100 80GB GPU, reducing wall-clock time 
by 1.7$\times$ relative to single LoRA (10h 8min 
vs 17h 0min) and 2.0$\times$ relative to full 
model fine-tuning (10h 8min vs 20h 22min). At 
inference time, parameter-free centroid routing 
adds negligible latency, with the base model 
remaining resident in GPU memory while only the 
active 52MB adapter is swapped per request. Full 
training and storage statistics are reported in 
Table~\ref{tab:efficiency}.

\begin{table}[H]
  \centering
  \tiny
  \begin{tabular}{lcc}
    \hline
    \textbf{Property} & \textbf{SemiAdapt-Instruct} & \textbf{Single LoRA} \\
    \hline
    Base model params   & 7,248M               & 7,248M          \\
    Adapter params      & 13.6M (0.19\%)       & 13.6M (0.19\%)  \\
    Params per domain   & 13.6M                & N/A             \\
    Total stored params & 884MB (17 adapters)  & 52MB            \\
    Parallel training   & 3--4 domains         & N/A             \\
    Initial training     & 10h 8min (3 processes) & 17h 0min  \\
Speedup vs single LoRA & 1.7x                       & 1x \\
Speedup vs full FT   & 2.0x                        & 1.2x \\
    New domain addition & 1 adapter only       & Full retrain    \\
    Domain update       & 1 adapter only       & Full retrain    \\
    \hline
  \end{tabular}
  \caption{Efficiency comparison between SemiAdapt-Instruct 
  and single LoRA fine-tuning.}
  \label{tab:efficiency}
\end{table}

\section{Conclusion}

SemiAdapt-Instruct establishes modular domain 
specialisation as a practical alternative to 
monolithic instruction tuning. By automatically 
discovering latent domains, training per-domain 
LoRA adapters, and parameter-free routing at 
inference, the framework matches single LoRA 
fine-tuning and outperforms full model 
fine-tuning across all configurations. We 
further demonstrate the extensibility of the framework 
by retraining a single adapter on 
new domain data and showing it outperforms all 
monolithic baselines. Together, these findings demonstrate that modular domain specialisation is a practical and extensible alternative to full retraining 
in evolving deployment settings.

\section*{Limitations}

SemiAdapt-Instruct is evaluated on a single 
heterogeneous English instruction dataset using 
Mistral-7B-Instruct-v0.3 as the base model. 
Whether the observed specialisation patterns 
generalise across languages, dataset compositions, 
or model families remains an open question, 
though cross-method convergence across three 
independent discovery mechanisms provides 
evidence that findings reflect instruction data 
structure rather than method or model artefacts.

ROUGE-L provides a universal comparison metric 
across heterogeneous domains but may underestimate 
gains for structured output domains where semantic 
correctness is not fully captured by lexical 
overlap. LLM-as-a-judge evaluation partially 
mitigates this limitation, though domain-specific 
evaluation metrics for automatically discovered 
domains remain an avenue for future work.

We acknowledge that comparison with learned routing approaches such as SLIM ~\cite{han-etal-2025-slim} and GLIDER represents an important direction for future work. The contribution of parameter-free centroid routing is not that it outperforms learned routers, but that it eliminates a trained component that would otherwise require maintenance and retraining as new domains are added. A learned router is structurally coupled to the domain set it was trained on. Centroid routing extends to new domains by recomputing a single mean embedding, with no retraining of any component.

\section{Acknowledgements}
This publication has emanated from research conducted with the financial support of Taighde Éireann - Research
Ireland under Grant number 18/CRT/6223.

\bibliography{custom}

\appendix

\label{sec:appendix}


\section{Data description}
\label{sec:datadescription}
We construct a heterogeneous instruction-tuning dataset by combining 
multiple publicly available corpora spanning diverse domains and task 
formats. All datasets are used for research purposes consistent with 
their respective licenses. Our mixture includes domain-specific datasets 
such as Finance-Alpaca\footnote{\url{https://huggingface.co/datasets/gbharti/finance-alpaca} (MIT License)} (financial instructions) and 
AlpaCare-MedInstruct\footnote{\url{https://huggingface.co/datasets/lavita/AlpaCare-MedInstruct-52k} (MIT License)} (medical question 
answering), alongside general-purpose and reasoning-focused resources. 
In particular, we incorporate WebGPT~\cite{nakano2022webgpt}, a 
retrieval-based question answering dataset (no explicit license; 
publicly available for research use), and Chain-of-Thought\footnote{\url{https://github.com/google-research/FLAN} (Apache 2.0)} (CoT) 
datasets from FLAN to enhance multi-step reasoning capabilities. To 
capture real-world problem-solving and conversational patterns, we 
further include StackOverflow dialogues~\cite{xu2023baize} (GPL-3.0).

To improve generalisation and coverage, we augment this mixture with 
additional general and code-oriented instruction datasets, namely 
Alpaca-cleaned\footnote{\url{https://huggingface.co/datasets/yahma/alpaca-cleaned} (CC BY-NC 4.0)}, Databricks 
Dolly\footnote{\url{https://huggingface.co/datasets/databricks/databricks-dolly-15k} (CC BY-SA 3.0)}, and 
CodeAlpaca\footnote{\url{https://huggingface.co/datasets/sahil2801/CodeAlpaca-20k} (CC BY 4.0)}. Overall, the resulting dataset balances 
domain-specific expertise (finance, medicine, programming) with general 
instruction-following and reasoning abilities, providing a heterogeneous 
dataset for research training. All trained artifacts derived from this 
data are intended for research use only, in accordance with the 
non-commercial restrictions of several constituent datasets (CC BY-NC 4.0).

StackOverflow Dialogues contributes 207,702 
examples (40.3\% of training data), 
substantially more than other sources. This 
imbalance influences K-means cluster boundaries: 
five of seventeen discovered clusters correspond 
to programming-related domains (c1, c3, c9, 
c13, c14), reflecting the dominance of 
StackOverflow content. While this may distort 
centroid positions toward programming embedding 
space, the high routing accuracy (95.0\%) and 
strong per-domain adapter performance for 
non-programming domains suggest the remaining 
clusters are sufficiently well-separated for 
effective specialisation. Future work should 
examine whether more balanced dataset 
composition affects domain discovery quality. Dataset composition and token statistics are 
summarised in Table~\ref{table:dataset} in 
the main paper.

\section{Soft Adapter Merging Ablation}

Centroid routing analysis reveals that 21.4\% 
of test examples exhibit routing ambiguity, 
defined as a cosine similarity gap below 0.05 
between the nearest and second-nearest domain 
centroids, indicating that a considerable 
subset of instructions are semantically 
related to multiple domains. Winning K-means 
domains show considerably lower ambiguity 
rates than losing domains: c10 (Mathematical 
Reasoning) and c11 (Natural Language 
Inference) exhibit ambiguity rates of 4.3\% 
and 2.8\% respectively, while adapter-losing 
domains such as c14 (48.4\%) and c4 (38.5\%) 
show substantially higher rates. This inverse 
relationship between routing confidence and 
specialisation benefit implies that domain 
coherence, manifesting in well-separated 
centroid representations, is necessary for 
optimal adapter specialisation.

Given the routing ambiguity observed for a 
subset of examples, we investigate whether 
soft adapter merging can improve performance 
for ambiguous examples. We implement this by 
computing a weighted average of the top-2 
adapter LoRA parameters directly, using 
normalised centroid similarity scores as 
weights. Contrary to expectation, soft merging 
consistently underperforms hard routing at 
both ambiguity thresholds tested: gap $<0.05$ 
($n=107$, $\Delta=-0.006$) and gap $<0.15$ 
($n=289$, $\Delta=-0.001$). At gap $<0.05$, similarity scores are nearly 
equal, producing approximately 50/50 weight 
averaging, which dilutes domain-specific 
specialisation without compensating benefit 
from the secondary adapter. These results confirm 
that hard centroid routing is the preferred 
inference strategy, with its 95\% oracle 
retention rate demonstrating near-optimal 
performance without merging overhead. The failure of uniform soft merging does not 
preclude more sophisticated combination 
strategies, though learned weighting would 
reintroduce a trained component, trading the 
extensibility advantage of parameter-free 
routing for potential quality gains on 
ambiguous examples.
\clearpage
\onecolumn

\section{Extended Per-Domain Analysis}

\label{sec:extendanalysis}

\begin{figure*}[h]
  \includegraphics[width=\columnwidth]{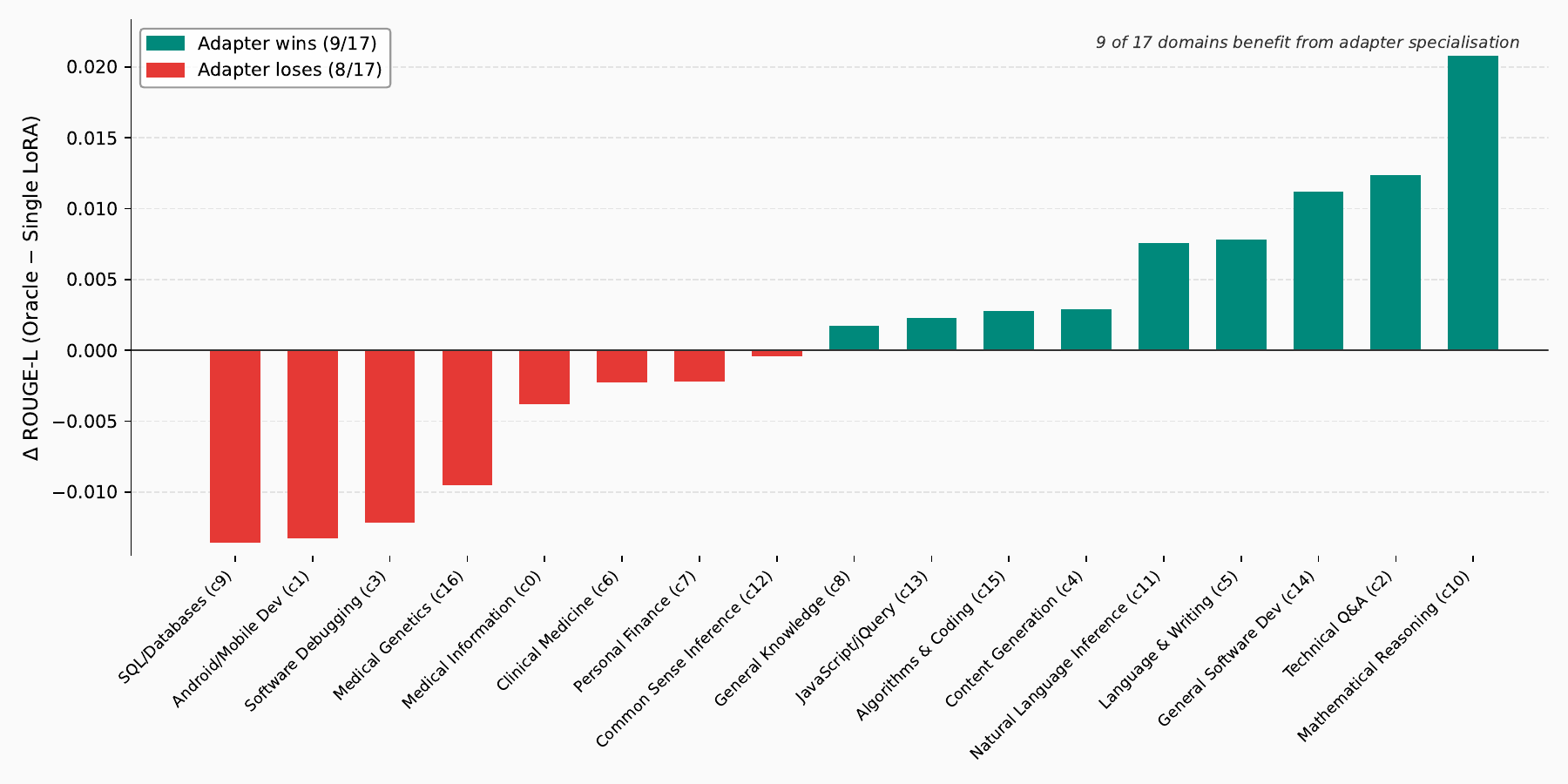}
  \caption{Visualisation of per-domain oracle ROUGE-L evaluation for K-means ($K$=17).}
  \label{fig:kmeans-oracle}
\end{figure*}

\begin{figure*}
  \includegraphics[width=\columnwidth]{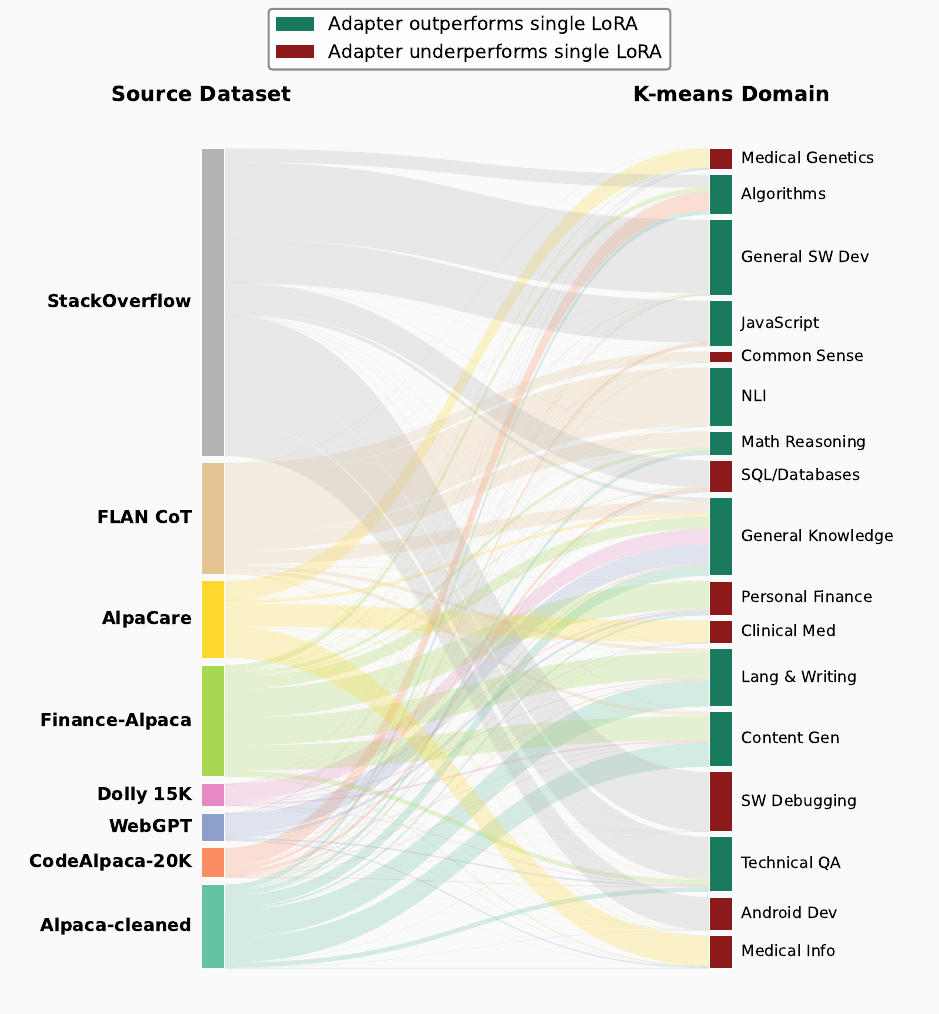}
  \caption{Source dataset composition across 
K-means ($K$=17) domain clusters. Flow width 
is proportional to the number of examples 
assigned to each cluster. Domain colours 
indicate adapter-winning (green) and 
adapter-losing (red) clusters under 
ROUGE-L evaluation.}
  \label{fig:kmeans-overlap}
\end{figure*}

\begin{figure*}[t]
  \includegraphics[width=\columnwidth]{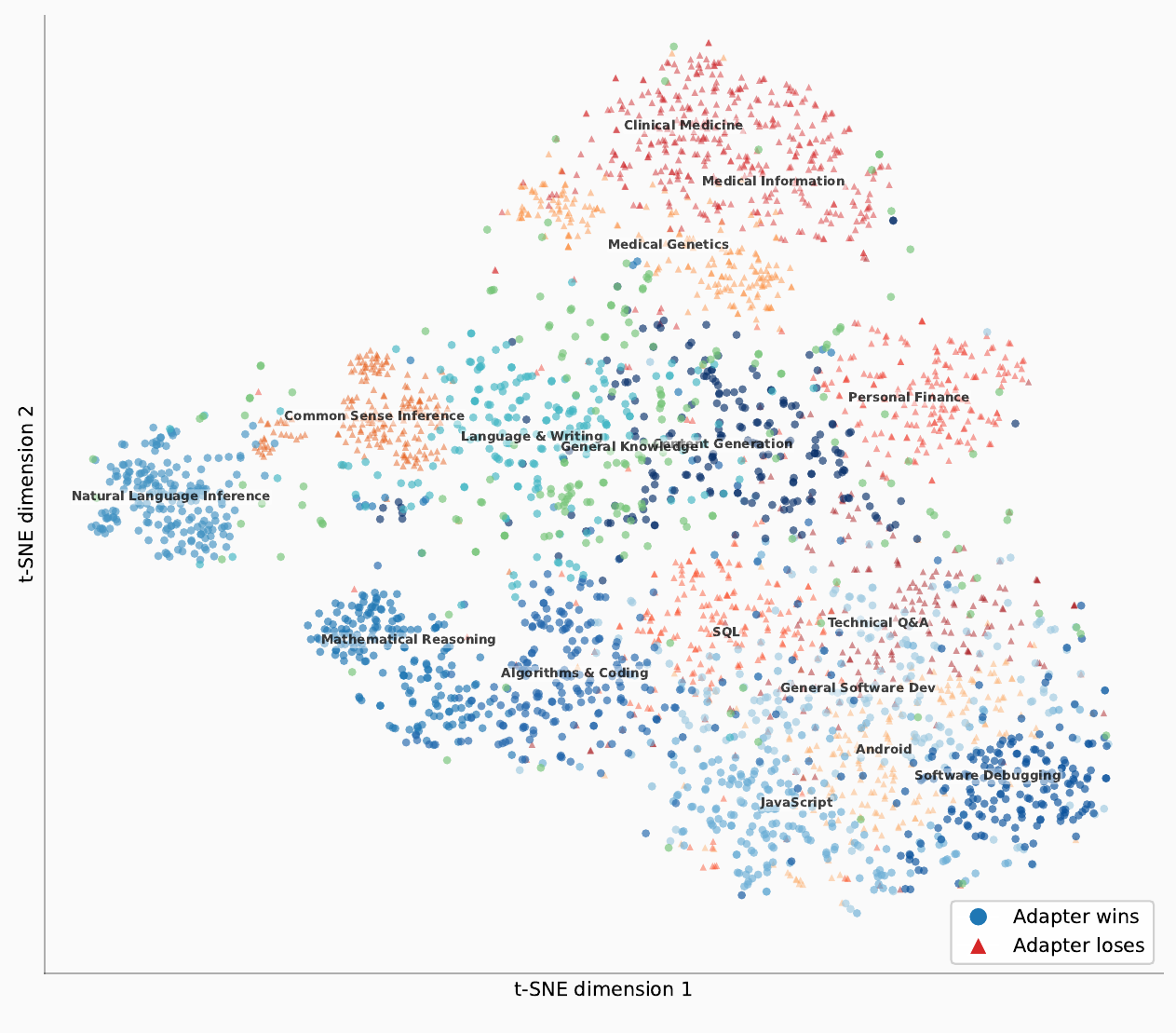}
  \caption{t-SNE projection of K-means ($K$=17) 
domain cluster assignments across the training 
set. Circles indicate adapter-winning clusters and triangles indicate adapter-losing clusters under ROUGE-L evaluation. }
  \label{fig:tsne}
\end{figure*}

\begin{table*}
\centering
\tiny
\begin{tabular}{rlp{4cm}rlp{4cm}}
\hline
\textbf{ID} & \textbf{N} & \textbf{Name} & \textbf{ID} & \textbf{N} & \textbf{Name} \\
\hline
$-$1 & 219,139 & -1\_human\_ai\_topic\_use              & 53 & 1,686 & 53\_cancer\_chemotherapy\_treatment\_breast \\
0  & 16,064 & 0\_sql\_table\_query\_database            & 54 & 1,680 & 54\_surgery\_knee\_pain\_bone \\
1  & 15,078 & 1\_title\_context\_com context\_com       & 55 & 1,667 & 55\_customer\_email\_customer service\_service \\
2  & 14,880 & 2\_step\_possible tell\_yes possible       & 56 & 1,610 & 56\_aws\_azure\_s3\_bucket \\
3  & 11,906 & 3\_stock\_money\_tax\_credit               & 57 & 1,598 & 57\_xml\_xml file\_xslt\_xml document \\
4  & 11,339 & 4\_states\_war\_title\_context             & 58 & 1,576 & 58\_numbers\_sum\_10\_print \\
5  & 10,808 & 5\_sentence\_following sentence\_rewrite   & 59 & 1,523 & 59\_learning\_machine learning\_neural \\
6  &  7,979 & 6\_think\_carefully make\_step             & 60 & 1,496 & 60\_certificate\_ssl\_encryption\_key \\
7  &  7,289 & 7\_claim\_noptions yes\_following          & 61 & 1,473 & 61\_disorder\_depression\_anxiety\_stress \\
8  &  7,163 & 8\_covid\_covid 19\_19\_outbreak           & 62 & 1,443 & 62\_mei ok\_mei\_leo\_ok let \\
9  &  6,356 & 9\_view\_android\_layout\_listview         & 63 & 1,356 & 63\_triangle\_area\_cm\_calculate area \\
10 &  6,313 & 10\_premise\_hypothesis\_step              & 64 & 1,317 & 64\_maps\_google maps\_google\_map \\
11 &  5,990 & 11\_animals\_species\_dogs\_animal         & 65 & 1,317 & 65\_spring\_spring boot\_boot\_bean \\
12 &  5,568 & 12\_string\_regex\_regular expression      & 66 & 1,243 & 66\_react\_component\_state\_props \\
13 &  5,293 & 13\_genetic\_gene\_mutation\_disease       & 67 & 1,198 & 67\_healthy\_exercise\_diet\_health \\
14 &  5,085 & 14\_request\_socket\_authentication\_http  & 68 & 1,187 & 68\_django\_model\_topic django\_field \\
15 &  4,962 & 15\_reasoning\_noptions\_reasoning process & 69 & 1,180 & 69\_sort\_sorting\_array\_order \\
16 &  4,737 & 16\_sentence\_sense noptions\_make sense   & 70 & 1,113 & 70\_thread\_threads\_mutex\_threading \\
17 &  4,525 & 17\_drug\_medication\_hypertension\_patient & 71 & 1,104 & 71\_random\_generate random\_random number \\
18 &  4,306 & 18\_date\_datetime\_time\_format            & 72 & 1,069 & 72\_opencv\_image\_cv\_images \\
19 &  4,234 & 19\_android\_app\_gradle\_android studio   & 73 & 1,033 & 73\_gridview\_datagrid\_row\_wpf \\
20 &  3,650 & 20\_class\_pointer\_type\_function         & 74 & 1,020 & 74\_song\_album\_band\_released \\
21 &  3,453 & 21\_angular\_angularjs\_ng\_directive       & 75 & 1,011 & 75\_qt\_creator\_topic qt\_pyqt \\
22 &  3,247 & 22\_directory\_command\_path\_file          & 76 & 1,007 & 76\_spark\_hadoop\_pyspark\_dataframe \\
23 &  3,145 & 23\_product\_marketing\_business\_iphone   & 77 & 1,000 & 77\_error\_error message\_message\_type \\
24 &  2,767 & 24\_plot\_chart\_matplotlib\_axis           & 78 &   995 & 78\_internet\_reddit\_title\_com \\
25 &  2,743 & 25\_patient\_year old\_symptoms\_old        & 79 &   978 & 79\_facebook\_api\_twitter\_app \\
26 &  2,725 & 26\_test natural\_language inference       & 80 &   951 & 80\_computer\_software\_cpu\_processor \\
27 &  2,698 & 27\_maven\_jar\_eclipse\_java               & 81 &   945 & 81\_headaches\_patient\_headache\_symptoms \\
28 &  2,689 & 28\_json\_object\_array\_json object        & 82 &   902 & 82\_think student\_teacher let\_premise \\
29 &  2,685 & 29\_team\_league\_football\_sports           & 83 &   889 & 83\_heart\_blood\_circulatory\_human heart \\
30 &  2,640 & 30\_div\_css\_height\_width                 & 84 &   886 & 84\_email\_mail\_outlook\_emails \\
31 &  2,432 & 31\_laravel\_topic laravel\_codeigniter     & 85 &   873 & 85\_stream consciousness\_random thoughts \\
32 &  2,420 & 32\_artificial intelligence\_robots         & 86 &   873 & 86\_maximum\_array\_maximum value \\
33 &  2,413 & 33\_visual studio\_visual\_studio\_dll      & 87 &   805 & 87\_statement\_statements\_loop\_boolean \\
34 &  2,273 & 34\_poem\_sky\_words\_sun                   & 88 &   794 & 88\_technology\_article\_impact\_essay \\
35 &  2,245 & 35\_climate\_climate change\_global warming & 89 &   775 & 89\_pdf\_pdf file\_document\_pdf files \\
36 &  2,190 & 36\_git\_branch\_repository\_commit         & 90 &   761 & 90\_rails\_gem\_ruby\_rake \\
37 &  2,187 & 37\_story\_short story\_generate\_character & 91 &   748 & 91\_website\_search\_web\_search engine \\
38 &  2,043 & 38\_diabetes\_type diabetes\_insulin        & 92 &   735 & 92\_firebase\_analytics\_authentication \\
39 &  2,033 & 39\_video\_audio\_play\_youtube              & 93 &   710 & 93\_decimal\_round\_rounding\_number \\
40 &  1,995 & 40\_heart\_chest\_chest pain\_myocardial    & 94 &   709 & 94\_array\_arr\_given array\_list \\
41 &  1,972 & 41\_dataframe\_pandas\_column\_df           & 95 &   665 & 95\_average\_deviation\_standard deviation \\
42 &  1,955 & 42\_lung\_chest\_patient\_cough              & 96 &   645 & 96\_vue\_vue js\_component\_js \\
43 &  1,924 & 43\_equation\_2x\_3x\_calculate             & 97 &   616 & 97\_alzheimer\_alzheimer disease\_amyloid \\
44 &  1,880 & 44\_le ok\_denny asked\_ok answer           & 98 &   613 & 98\_salary\_employees\_sql query \\
45 &  1,798 & 45\_mei ok\_mei\_leo\_ok let                & 99 &   602 & 99\_class\_create class\_student\_grade \\
46 &  1,785 & 46\_medical\_medical school\_education      & 100 &  577 & 100\_tensorflow\_tensor\_model\_keras \\
47 &  1,774 & 47\_nonsensical\_following sentences        & 101 &  567 & 101\_webpack\_babel\_module\_npm \\
48 &  1,770 & 48\_le ok\_denny asked\_ok answer           & 102 &  564 & 102\_ethical\_medical\_ethics\_patient \\
49 &  1,761 & 49\_series\_film\_consciousness             & 103 &  551 & 103\_utf\_file\_encoding\_text file \\
50 &  1,750 & 50\_jquery\_element\_div\_javascript        & 104 &  534 & 104\_dog\_premise\_hypothesis entailed \\
51 &  1,721 & 51\_docker\_jenkins\_container\_kubernetes  & 105 &  513 & 105\_prime\_prime numbers\_prime number \\
52 &  1,714 & 52\_mongodb\_elasticsearch\_mongoose        &     &       & \\
\hline
\end{tabular}
\caption{BERTopic 107-topic domain labels and training set counts.}
\label{tab:bertopic107_labels}
\end{table*}

\begin{table*}[h]
\centering
\small
\begin{tabular}{cp{9cm}}
\hline
\textbf{Super-24 Topic} & \textbf{BERTopic-107 Topic IDs} \\
\hline
super\_01 & 71, 105, 58, 95, 43, 63, 18, 93 \\
super\_02 & 94, 86, 69 \\
super\_03 & 75, 73, 9, 19, 39, 64, 92, 52, 28 \\
super\_04 & 50, 30 \\
super\_05 & 59, 32 \\
super\_06 & 47, 16, 11, 1, 15, 29, 4 \\
super\_07 & 80, 91, 3, 55, 78, 23, 88, 35 \\
super\_08 & 67, 38, 25, 13, 81, 61, 97 \\
super\_09 & 102, 46 \\
super\_10 & 101, 66, 21, 96 \\
super\_11 & 36, 51, 56, 76 \\
super\_12 & 87, 99, 20, 70 \\
super\_13 & 41, 24 \\
super\_14 & 6, 85 \\
super\_15 & 62, 45 \\
super\_16 & 60, 79, 14, 84 \\
super\_17 & 53, 17, 54, 83, 40, 42, 8 \\
super\_18 & 33, 77, 22, 90, 31, 12, 103, 89, 57 \\
super\_19 & 65, 27 \\
super\_20 & 100, 68, 72 \\
super\_21 & 104, 26, 10, 82, 2, 7 \\
super\_22 & 0, 98 \\
super\_23 & 44, 48 \\
super\_24 & 49, 74, 37, 34, 5 \\
\hline
\end{tabular}
\caption{BERTopic super-24 groupings showing constituent 
BERTopic-107 topic IDs. Super-groups were derived by 
merging semantically related topics via intertopic 
distance mapping.}
\label{tab:super24_groupings}
\end{table*}

\end{document}